\documentclass[10pt,twocolumn,letterpaper]{article}

\usepackage{cvpr}
\usepackage{times}
\usepackage{epsfig}
\usepackage{graphicx}
\usepackage{amsmath}
\usepackage{amssymb}
\usepackage{algorithm}
\usepackage{algpseudocode}
\DeclareMathOperator*{\argmin}{argmin}   

\usepackage[hyphens]{url}
\usepackage[flushleft]{threeparttable}
\usepackage{tablefootnote}
\usepackage[]{subfig}

\usepackage[pagebackref=true,breaklinks=true,letterpaper=true,colorlinks,bookmarks=false]{hyperref}

\cvprfinalcopy 

\def\cvprPaperID{6821} 

\ifcvprfinal\fi
\begin{document}

\title{Fingerprints: Fixed Length Representation via \\Deep Networks and Domain Knowledge}

\author{Joshua J. Engelsma, Kai Cao, Anil K. Jain\\
Michigan State University\\
East Lansing, MI, USA\\
{\tt\small \{engelsm7, kaicao, jain\}@cse.msu.edu}
}

\maketitle

\begin{abstract}
   We learn a discriminative fixed length feature representation of fingerprints which stands in contrast to commonly used unordered, variable length sets of minutiae points. To arrive at this fixed length representation, we embed fingerprint domain knowledge into a multitask deep convolutional neural network architecture. Empirical results, on two public-domain fingerprint databases (NIST SD4 and FVC 2004 DB1) show that compared to minutiae representations, extracted by two state-of-the-art commercial matchers  (Verifinger v6.3 and Innovatrics v2.0.3), our fixed-length representations provide (i) higher search accuracy: Rank-1 accuracy of 97.9\%~vs.~97.3\% on NIST SD4 against a gallery size of 2000 and (ii) significantly faster, large scale search: 682,594 matches per second vs. 22 matches per second for commercial matchers on an i5 3.3 GHz processor with 8 GB of RAM.
\end{abstract}

\section{Introduction}
\begin{figure}[t]
  \centering
  \subfloat[Level-1 features]{\includegraphics[scale=0.52]{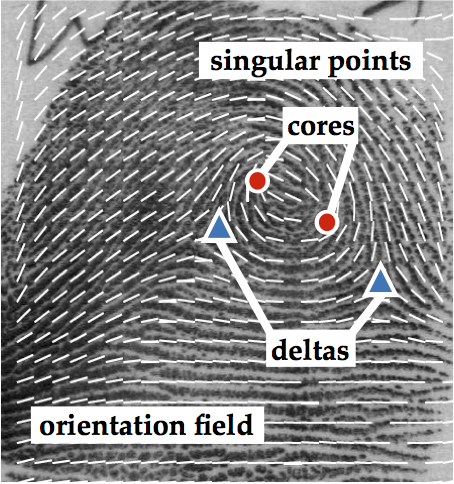}\label{fig:f1}}
  \hfill
  \subfloat[Level-2 features]{\includegraphics[scale=0.52]{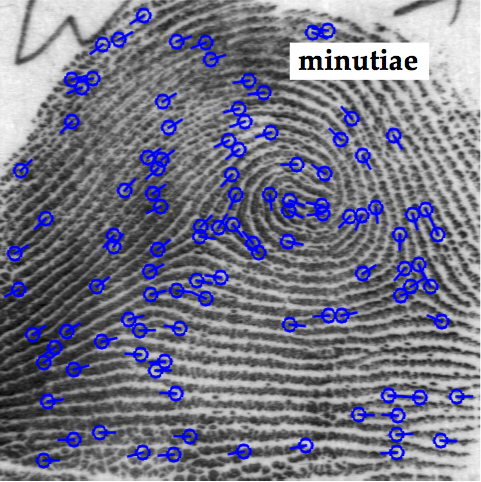}\label{fig:f2}}
  \caption{Traditional fingerprint representations consist of global level-1 features (a) and local level-2 features, called minutiae points, together with their descriptors (b).}
  \label{fig:intro}
\vspace{-1.0em}
\end{figure}

Over 100 years ago, the pioneering giant of modern day fingerprint recognition, Sir Francis Galton, astutely commented on fingerprints in his 1892 book titled ``Finger Prints":
\begin{quote}
\textit{``They have the unique merit of retaining all their peculiarities unchanged throughout life, and afford in consequence an incomparably surer criterion of identity than any other bodily feature."}~\cite{galton}
\end{quote}
Galton went on to describe fingerprint \textit{minutiae}, the small details woven throughout the papillary ridges on each of our fingers, which Galton believed provided uniqueness and permanence properties for accurately identifying individuals. In the 100 years since Galton's ground breaking scientific observations, fingerprint recognition systems have become ubiquitous and can be found in a plethora of different domains~\cite{handbook} such as forensics~\cite{ngi}, healthcare, mobile device security~\cite{touchid},  mobile payments~\cite{touchid}, border crossing~\cite{obim}, and national ID~\cite{india1}. To date, virtually all of these systems continue to rely upon the location and orientation of minutiae within fingerprint images for recognition (Fig.~\ref{fig:intro}).

\newcommand{\specialcell}[2][c]{%
  \begin{tabular}[#1]{@{}c@{}}#2\end{tabular}}
  \newcommand{\tabitem}{~~\llap{\textbullet}~~}

Although fingerprint recognition systems based on minutiae (i.e. handcrafted features) have enjoyed significant success over the years, to our knowledge, not much effort has been devoted to augment handcrafted features with recent advances in deep learning to improve the recognition accuracy and reduce the complexity of large scale search. The argument for introducing deep learning models to fingerprint recognition is compelling given the following major limitations of prevailing minutiae-based fingerprint recognition systems.

\begin{enumerate}
\item Minutiae-based representations are of variable length (Table~\ref{table:templates}), since the number of extracted minutiae varies amongst different fingerprint images even of the same finger (Fig.~\ref{fig:intro_fig} (a)). This causes two main problems: (i) pairwise fingerprint comparison is computationally demanding, accentuated when searching databases with an inordinate number of identities, e.g., India's Aadhaar system with 1.25 billion identities~\cite{india1} and (ii) template encryption, a necessity for user privacy protection, is a significant challenge~\cite{encryption}.

\begin{figure}[t]
\begin{center}
\includegraphics[scale=0.5]{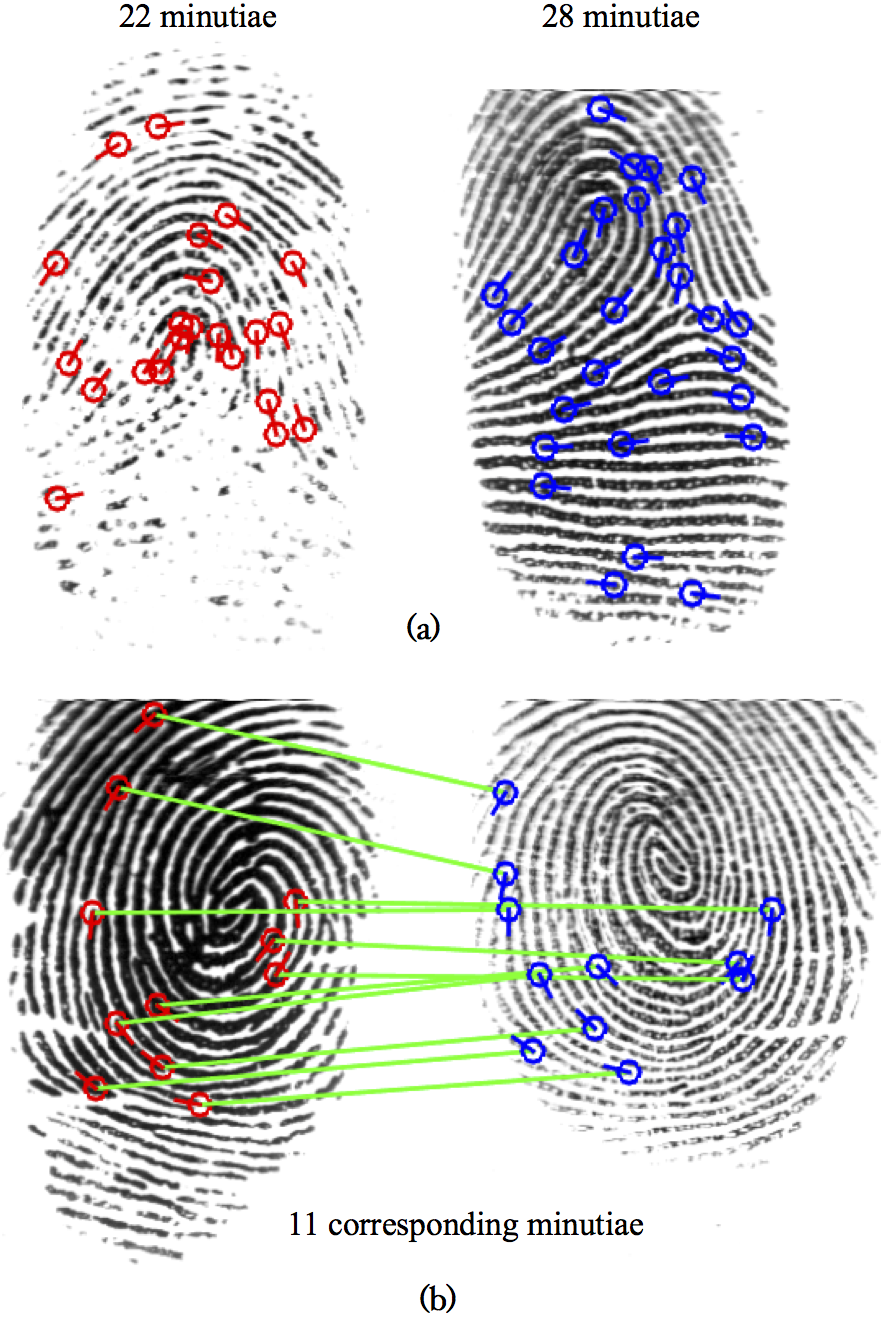}
\caption{Failures of a minutiae-based matcher (COTS A). The genuine pair (two impressions from the same finger) in (a) was falsely rejected at 0.1\% FAR due to inaccurate minutiae extraction. The imposter pair (impressions from two different fingers) in (b) was falsely accepted at 0.1\% FAR due to the similar minutiae distribution in these two fingerprints.}
\label{fig:intro_fig}
\end{center}
\vspace{-1.0em}
\end{figure} 

\begin{figure}[t]
\begin{center}
\includegraphics[scale=0.4]{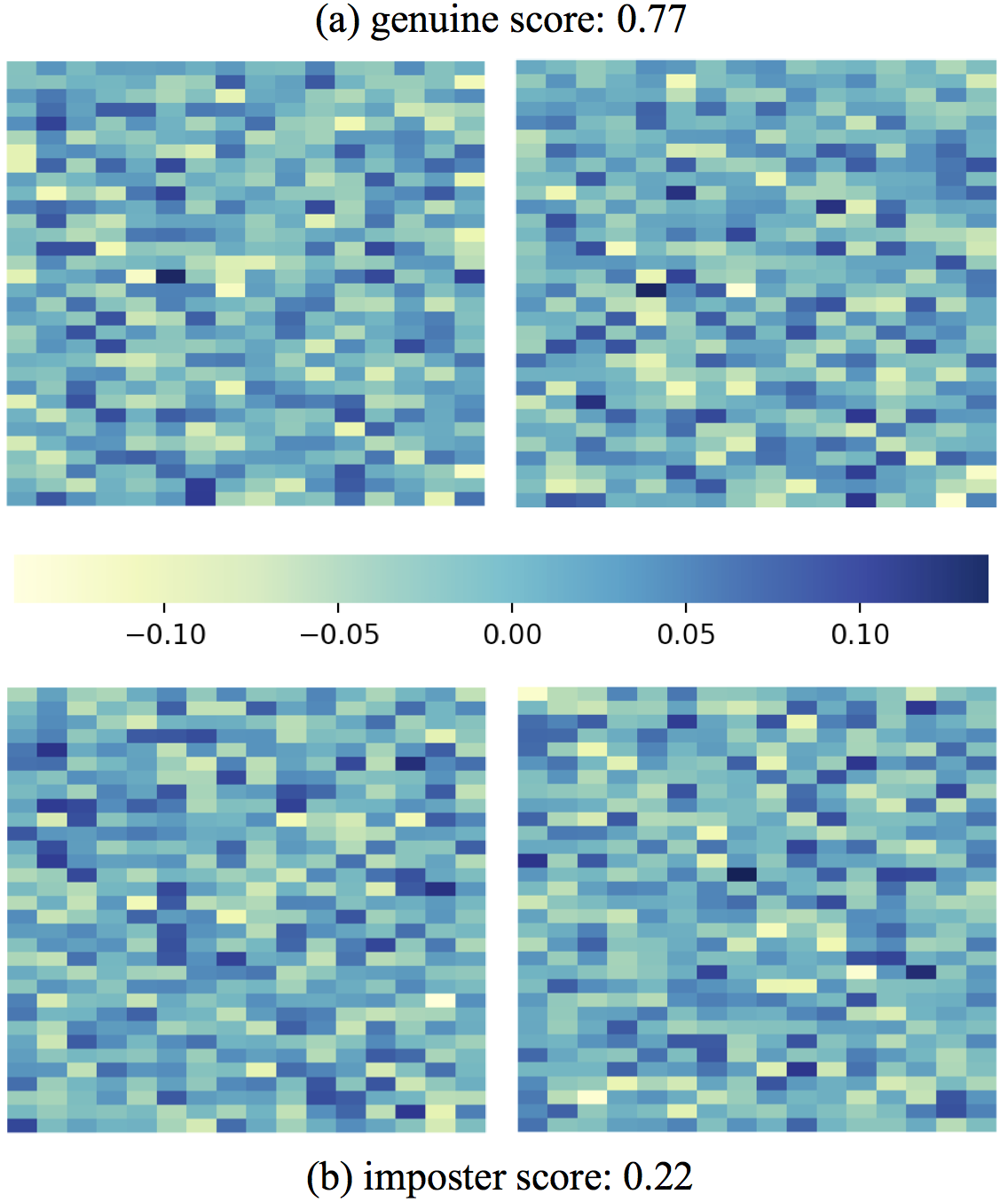}
\caption{Proposed fixed-length, 512-dimensional fingerprint representations extracted from the same four fingerprints shown in Figure~\ref{fig:intro_fig}. Unlike COTS A, we correctly classify the pair in (a) as a genuine pair, and the pair in (b) as an imposter pair. The score threshold of our matcher @ FAR = 0.1\% is 0.69}
\label{fig:feature_maps}
\end{center}
\vspace{-1.0em}
\end{figure} 

\item Fingerprint recognition can be viewed as a \textbf{75 billion class problem} ($\approx$~7.5 billion living people, each with 10 fingers) with large intra-class variability and large inter-class similarity (Fig.~\ref{fig:intro_fig}). This necessitates devising extremely discriminative yet compact representations that go beyond just minutiae points.

\item Reliable minutiae extraction in low quality fingerprints (due to noise, distortion, finger condition) is problematic, causing false rejects in the recognition system (Fig.~\ref{fig:intro_fig} (a)). See also NIST fingerprint evaluation FpVTE 2012~\cite{nist}.
\end{enumerate}

\begin{table}[h]
\small
\caption{Template Comparisons\tnote{1}}
 \centering
\begin{threeparttable}
\begin{tabular}{|c|c|c|}
 \hline
 Matcher & \specialcell{(Min, Max) \\ \# of Minutiae}  & \specialcell{(Min, Max) \\Template Size (kB)} \\
 \hline
 \specialcell{COTS A} & (8, 206) & (1.1, 22.6) \\
  \hline
 \specialcell{COTS B} & (6, 221) & (0.1, 1.3) \\
  \hline
  \hline
   \specialcell{Proposed} & N.A.\tnote{2} & 2\tnote{\textdagger} \\
  \hline
\end{tabular}
\begin{tablenotes}
\item[1] Statistics from NIST SD4 and FVC 2004 DB1.
\item[2] Template is not explicitly comprised of minutiae.
\item[\textdagger] Template size is fixed at 2 kilobytes, irrespective of the number of minutiae.
\end{tablenotes}
\end{threeparttable}
\label{table:templates}
\vspace{-1.em}
\end{table}

\begin{figure*}[t]
\begin{center}
\includegraphics[scale=0.5]{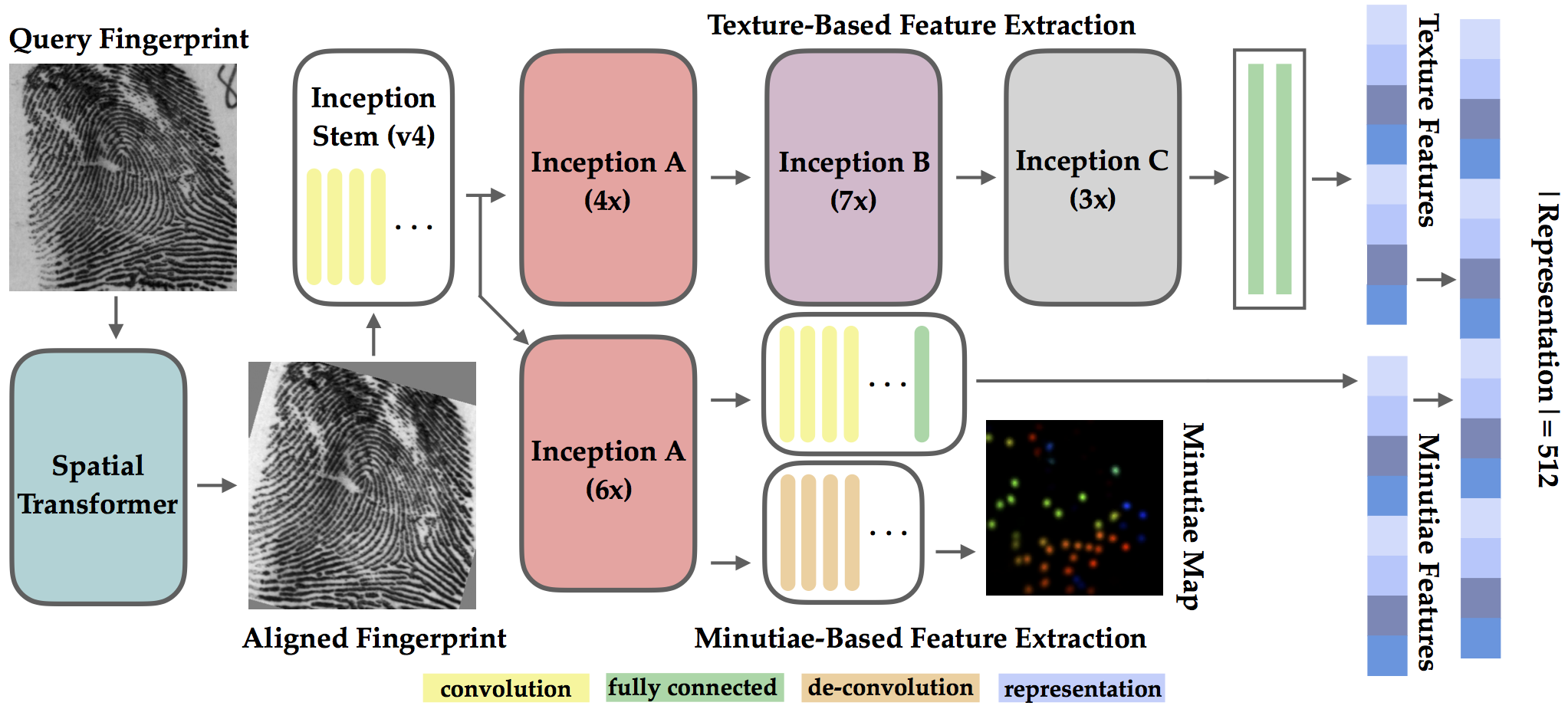}
\caption{Flow diagram of the proposed deep learning based fingerprint recognition system: (i) a query fingerprint is cropped and aligned via the spatial transformer; (ii) the aligned fingerprint is passed to the Inception v4 stem which extracts both a textural representation and a minutiae-map based representation; (iii) The texture representation and minutiae representation are concatenated into a 512-dimensional compact representation of the input fingerprint.}
\label{fig:schematic}
\end{center}
\vspace{-1.5em}
\end{figure*} 

Given the limitations of minutiae-based matchers, we propose a custom deep network to extract discriminative fixed-length representations (Fig.~\ref{fig:feature_maps}) for state-of-the-art fingerprint recognition and search. Unlike several prior attempts to arrive at a fixed length fingerprint representation~\cite{index1, index2}, our method incorporates fingerprint domain knowledge into the deep network architecture via a \textit{minutiae-map}. More specifically, our architecture learns both a \textit{texture based representation} and a \textit{minutiae-map based representation}. The concatenation of these two representations comprises our fixed-length fingerprint representation. In addition to injecting domain knowledge into the network, we add a spatial transformer module~\cite{spatial} to crop and align the query (probe) fingerprint prior to extracting its representation. Finally, we reduce the complexity of our network architecture and inference time via a variant of the teacher-student training paradigm.

The primary contributions of this work are as follows:

\begin{enumerate}

\item A custom deep network architecture that utilizes fingerprint domain knowledge (minutiae locations and orientations) to extract a discriminative fixed-length representation.

\item Empirical results demonstrating a significant improvement in fingerprint recognition speed (a \textbf{31,500 fold} improvement in the number of matches per second) over two fingerprint SDKs on two benchmark datasets (NIST SD4 and FVC 2004 DB1) while retaining comparable recognition accuracy.

\item A method for significantly reducing the memory consumption and inference time of the network using the teacher-student training paradigm. 

\end{enumerate}

\section{Prior Work}

While some attention has been given towards using deep learning models in fingerprint recognition, most of these works are focused on improving only a \textit{sub-module} of an end-to-end fingerprint recognition system such as segmentation~\cite{seg1, seg2, seg3, seg4}, orientation field estimation~\cite{orien1,orien2,orien3}, minutiae extraction~\cite{minut1, minut2, minut3}, and minutiae descriptor extraction~\cite{descript1}. While they are able to improve the performance of the sub-modules, they still operate within the conventional paradigm of extracting a variable length feature representation comprised of the minutiae and their associated descriptors in a given fingerprint image. 

We completely reformulate the conventional fingerprint recognition paradigm. Rather than working towards an end goal of minutiae extraction for matching, we allow a deep network to \textit{learn} a fixed-length feature vector that best discriminates fingerprints. While there have been two previous attempts along this direction~\cite{index1, index2}, both require computationally intensive global alignment of fingerprints prior to extracting fixed length representations via a black box deep network. The authors in~\cite{fingercode} also proposed a fixed length fingerprint representation based upon handcrafted textural features. Both~\cite{index1, index2} and~\cite{fingercode} have inferior accuracy compared to minutiae-based systems. In contrast, our domain knowledge infused multitask network with built in alignment (via the spatial transformer module~\cite{spatial}) is competitive with state-of-the-art minutiae-based matchers in accuracy while being significantly faster. The network is also more interpretable than~\cite{index1, index2} given the minutiae map output. The speed and accuracy of our algorithm will open up new opportunities for large scale search and matching.

\section{Approach}

We provide an overview and intuition of the proposed end-to-end, domain knowledge infused, deep learning based fingerprint recognition system. We then describe how incorporating a spatial transformer module into our network enables fingerprint cropping and alignment as part of the same network inference used for representation extraction. Finally, we discuss the domain knowledge (minutiae map) that is injected into the network via a multitask learning objective.

\begin{algorithm}
 
\caption{Extract Fingerprint Representation} \begin{algorithmic}[1]
  \State \textbf{$l(I)$}: shallow localization network
  \State \textbf{$g(I,x,y,\theta)$}: bilinear grid sampler
  \State \textbf{$s(I)$}: inception stem
  \State \textbf{$m(F_{map})$}: minutiae branch
  \State \textbf{$t(F_{map})$}: texture branch
\State 
\State \textbf{Input:} Unaligned $448\times448$ fingerprint image $I_f$
\State $(x, y, \theta) \gets l(I_f)$
\State $I_t \gets g(I_f, x, y, \theta)$
\State $F_{map} \gets s(I_t)$
\State $R_{minutiae} \gets m(F_{map})$
\State $R_{texture} \gets t(F_{map})$
\State $R_{fused} \gets R_{minutiae} \oplus R_{texture}$
\State \textbf{Output:} fingerprint representation $R_{fused} \in \mathbb{R}^{512}$
\end{algorithmic} 
\label{alg:alg1}
\end{algorithm} 

\subsection{Overview}

A high level overview of our proposed network architecture is provided in Figure~\ref{fig:schematic} with pseudocode describing the process shown in Algorithm~\ref{alg:alg1}. The model is trained with a longitudinal database comprised of 440K rolled fingerprint images stemming from 37,410 unique fingers~\cite{longitudinal}. The primary task during training is to predict the finger class label $c \in [0,37410]$ of each of the 440K training fingerprint images ($\approx12$ fingerprint impressions / finger). Similar to prevailing face recognition systems, the last fully connected layer can be used as the representation for fingerprint matching.

The input to our model is a 448x448 grayscale fingerprint image which is first passed through the spatial transformer module. The spatial transformer acts as the network's built-in fingerprint cropping and alignment module. After applying the spatial transformation to $I_f$, a cropped and aligned fingerprint $I_t$ is passed to the base network. 

The backbone of our base network is the Inception v4 architecture proposed in~\cite{inceptionv4}. We specifically modified the network architecture to incorporate two different branches (Fig.~\ref{fig:schematic}) using the three Inception modules (A, B, and C) described in~\cite{inceptionv4}. The first branch performs the primary learning task of predicting a finger class label $c$ directly from the cropped, aligned fingerprint $I_t$ and essentially learns the texture cues in the fingerprint image. The second branch again predicts the finger class label $c$ from the aligned fingerprint $I_t$, but it also has a related side task of predicting the minutiae locations and orientations in $I_t$. In this manner, we guide this branch of the network to extract representations influenced by fingerprint minutiae. The textural cues act as complementary discriminative information to the minutiae-guided representation. The final feature representation is the 512-dimensional concatenation of the two representations. Note that the minutiae set is not explicitly used in the final representation, we only use the minutiae-map to guide our network training.

In the following subsections, we provide details of the major subcomponents of the proposed network architecture.

\begin{figure}[h]
  \centering
  \subfloat[Input fingerprint]{\includegraphics[scale=0.32]{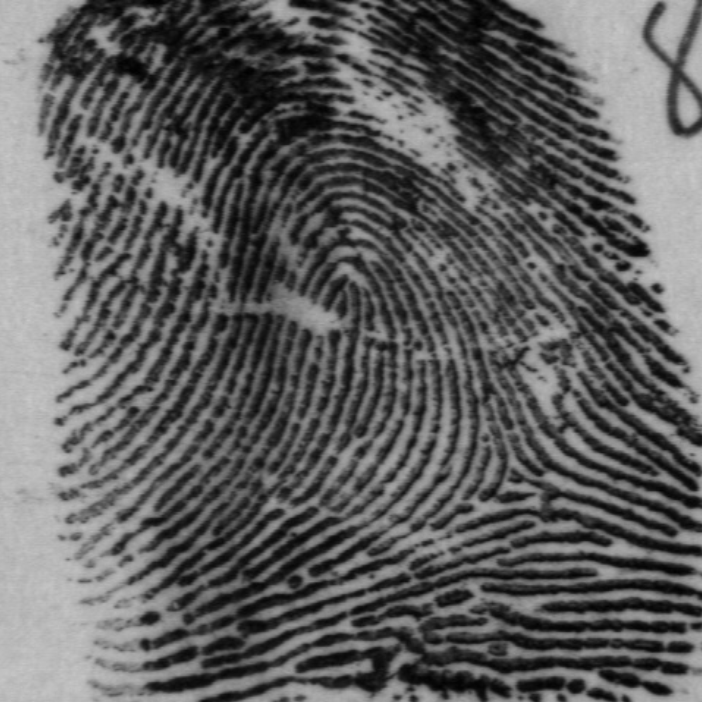}\label{fig:f1}}
  \hfill
  \subfloat[Aligned fingerprint]{\includegraphics[scale=0.32]{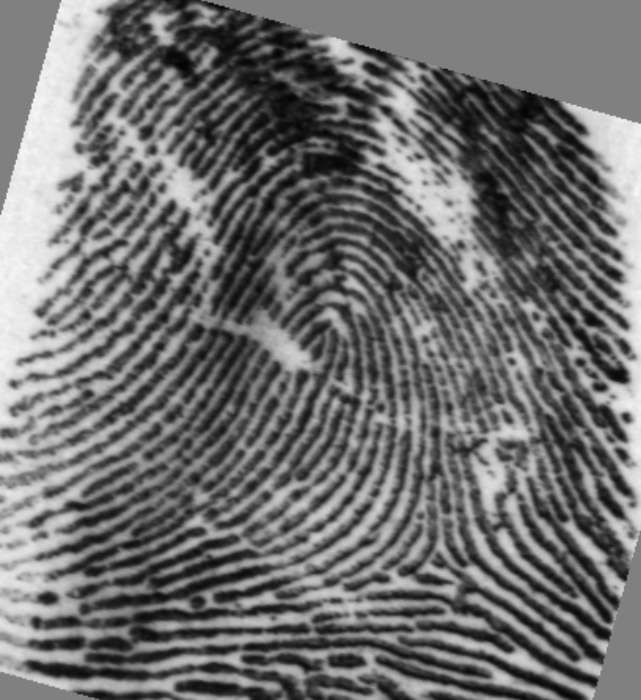}\label{fig:f2}}
  \caption{An unaligned fingerprint (a) is cropped and aligned (b) by the spatial transformer module prior to representation extraction.}
  \label{fig:transform}
\vspace{-1.0em}
\end{figure}

\subsection{Cropping and Alignment}

In nearly all minutiae-based fingerprint recognition systems, the first step is to crop a region of interest (ROI) and then perform global alignment based on some reference points (such as the core point). However, this alignment is computationally expensive. This motivated us to consider attention mechanisms such as the spatial transformers in~\cite{spatial}. 

The advantages of using the spatial transformer module in place of reference point based alignment algorithms are two-fold: (i) it requires only one forward pass through a shallow localization network (Table~\ref{tabel:localization}), followed by bilinear grid sampling. This reduces the computational complexity of alignment; (ii) The parameters of the localization network are tuned to minimize the classification loss of the base-network (representation extraction network). In other words, we let the base-network decide what a ``good" transformation is, so that it can better classify the input fingerprints.

\begin{table}[t]
\caption{Localization Network Architecture}
 \centering
\begin{threeparttable}
\begin{tabular}{|c c c|}
 \hline
 Type & \specialcell{Output \\ Size} & \specialcell{Filter \\ Size, Stride}\\
 \hline
 Convolution & $128\times128\times24$ & $5\times5$, $1$\\
 \hline
 Max Pooling & $64\times64\times24$ & $2\times2$, $2$\\
  \hline
 Convolution & $64\times64\times32$ & $3\times3$, $1$\\
  \hline
 Max Pooling & $32\times32\times32$ & $2\times2$, $2$\\
  \hline
 Convolution& $32\times32\times48$ & $3\times3$, $1$\\
  \hline
 Max Pooling & $16\times16\times48$ & $2\times2$, $2$\\
  \hline
 Convolution& $16\times16\times64$ & $3\times3$, $1$\\
  \hline
 Max Pooling & $8\times8\times64$ & $2\times2$, $2$\\
  \hline
 Fully Connected & $64$ &  \\
 \hline
  Fully Connected & $3$ &  \\
 \hline
\end{tabular}
\end{threeparttable}
\label{tabel:localization}
\end{table}

\begin{figure*}[t]
\begin{center}
\includegraphics[scale=0.5]{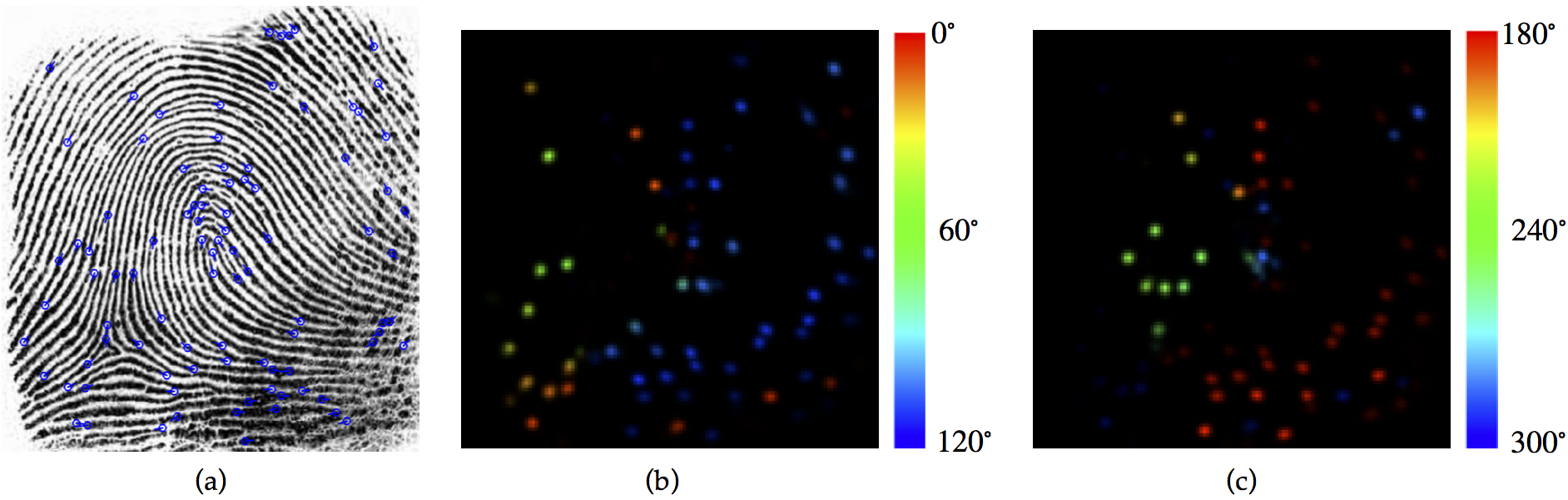}
\caption{Minutiae Map Extraction. The minutiae locations and orientations of an input fingerprint (a) are encoded as a 6-channel minutiae map. The ``hot spots" in each channel indicate the spatial location of the minutiae points. The color of the hot spots indicate their orientations.}
\label{fig:mmap}
\end{center}
\vspace{-1.0em}
\end{figure*} 

Given an uncropped, unaligned fingerprint image $I_f$, a shallow localization network first hypothesizes the parameters of an affine transformation matrix $A_\theta$. Given $A_\theta$, a grid sampler $T_\theta(G_i)$ samples the input image $I_f$ pixels $(x_i^f, y_i^f)$ for every target grid location $(x_i^t, y_i^t)$ to output the cropped, aligned fingerprint image $I_t$ in accordance with Equation 1.

\begin{equation}
\begin{pmatrix}x_i^f\\ y_i^f\end{pmatrix} = T_\theta(G_i)  = A_\theta\begin{pmatrix}x_i^t\\ y_i^t\\ 1\end{pmatrix}
\end{equation} Once $I_t$ has been computed from the grid sampler $T_\theta(G_i)$, it is passed on to the base network for classification. Finally, the parameters for the localization network are updated based upon the classification loss of the base network.

The architecture used for our localization network is shown in Table~\ref{tabel:localization} and images from before and after the spatial transformer module are shown in Figure~\ref{fig:transform}. In order to get the spatial transformer to properly converge, (i) the learning rate for the localization network parameters was scaled by $0.035$, (ii) the upper bound of the estimated affine matrix translation and rotation parameters was set to $224$ pixels and $\pm 60$ degrees, respectively, and (iii) the affine matrix scaling parameter was fixed to select a window of size $285\times285$ from the input fingerprint image. These constraints are based on our domain knowledge on the maximum extent a user would rotate / translate their fingers during placement on the reader platen and the typical ``high quality" area of a fingerprint image (usually the center portion).

\subsection{Minutiae Map Domain Knowledge}

To prevent overfitting the network to the training data, we incorporate fingerprint domain knowledge into the network to guide it to learn a more discriminative representation. The specific domain knowledge we incorporate into our network architecture is hereafter referred to as the \textit{minutiae map}~\cite{cao}. Note that the minutiae map is not explicitly used in the fingerprint representation, but the information contained in the map is indirectly embedded in the network during training.

A minutiae map is essentially a $c$-channel heatmap quantizing the locations $(x, y)$ and orientations $\theta \in [0,2\pi]$ of the minutiae within a fingerprint image. More formally, let $h$ and $w$ be the height and width of an input fingerprint image $I_f$ and $T = \{m_1, m_2, ..., m_n\}$ be its minutiae template with $n$ minutiae points, where $m_t = (x_t, y_t, \theta_t)$ and $t = 1,...,n$. Then, the minutiae map $H \in \mathbb{R}^{h\times w\times c}$ at $(i, j, k)$ can be computed by summing the location and orientation contributions of each of the minutiae in $T$. More formally,

\begin{equation}
H(i, j, k) = \sum_{t=1}^n C_s((x_t,y_t), (i, j)) \cdot C_o(\theta_t, 2k\pi /c)
\end{equation} where $C_s(.)$ and $C_o(.)$ calculate the spatial and orientation contribution of minutiae $m_t$ to the minutiae map at $(i, j, k)$ based upon the euclidean distance of $(x_t,y_t)$ to $(i,j)$ and the orientation difference between $\theta_t$ and $2k\pi /c$ as follows:

\begin{equation}
C_s((x_t,y_t), (i,j)) = exp(-\frac{||(x_t,y_t)-(i,j)||_2^2}{2\sigma_s^2})
\end{equation}

\begin{equation}
C_o(\theta_t,2k\pi /c) = exp(-\frac{d\phi(\theta_t,2k\pi /c)}{2\sigma_s^2})
\end{equation} where $d\phi(\theta_1,\theta_2)$ is the orientation difference between angles $\theta_1$ and $\theta_2$:

\begin{equation}
d\phi(\theta_1, \theta_2) = \begin{cases}
|\theta_1 - \theta_2| & -\pi \leq -\theta_1 - \theta_2 \leq \pi \\
2\pi - |\theta_1 - \theta_2| & otherwise.
\end{cases}
\end{equation} In our approach, we extract minutiae maps of size $128\times128\times6$ to encode the minutiae locations and orientations of an input fingerprint image of size $448\times448\times1$. An example fingerprint image and its corresponding minutiae map are shown in Figure~\ref{fig:mmap}. In the following subsection, we will explain in detail how we inject the information contained in the minutiae maps into our custom architecture.

\subsection{Multi-Task Architecture}

\begin{figure}[h]
\begin{center}
\includegraphics[scale=0.5]{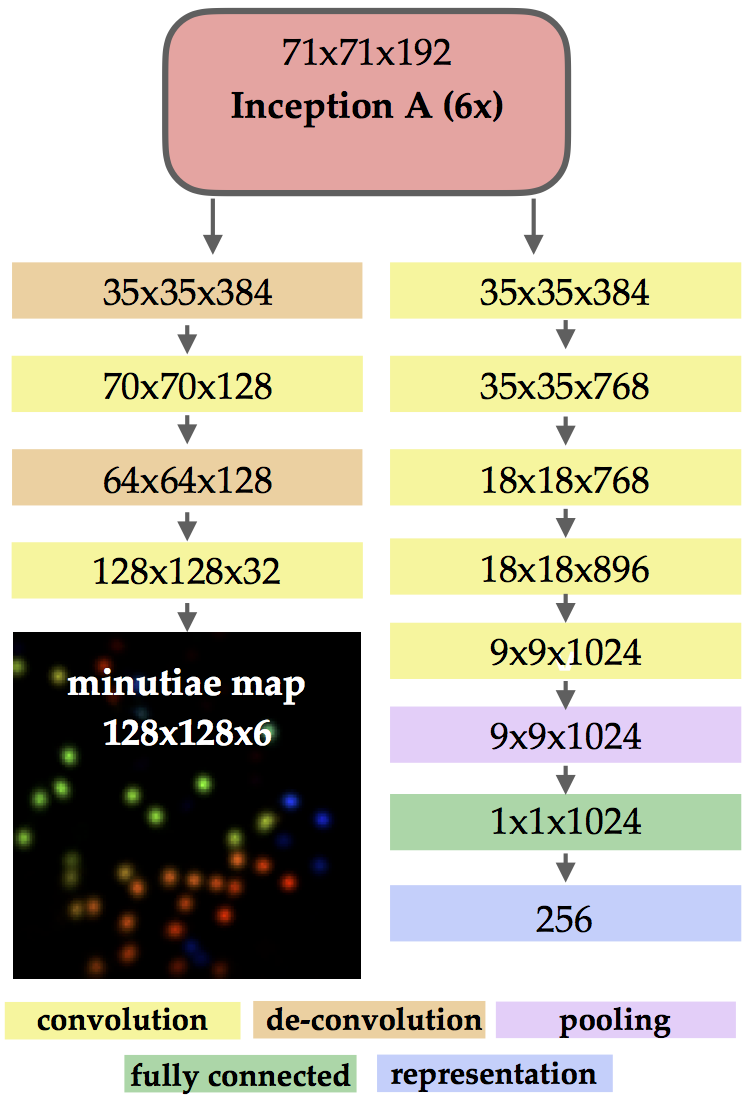}
\caption{The custom multi-task \textbf{minutiae branch} $m_{w_2}$ of our proposed network. The dimensions inside each box represent the input dimensions.}
\label{fig:minutiae_arch}
\end{center}
\vspace{-1.0em}
\end{figure}

The minutiae-map domain knowledge is injected into our Inception v4 network backbone via multitask learning. Multitask learning improves generalizability of a model since domain knowledge within the training signals of related tasks acts as an inductive bias~\cite{multi, multi0, multi1, multi2, multi3, multi4}. The multi-task branch of our network architecture is shown in Figures~\ref{fig:schematic} and~\ref{fig:minutiae_arch}. The primary task of the branch is to classify a given fingerprint image into its ``finger class" and the secondary task is to estimate the minutiae-map. In this manner, we guide the minutiae-branch of our network to extract fingerprint representations that are influenced by minutiae locations and orientations. A separate branch in our network aims to extract a complementary texture-based representation by directly predicting the class label of an input fingerprint without any domain knowledge (Fig.~\ref{fig:schematic}). 

Note, we combine the texture branch with the minutiae branch in our architecture (rather than two separate networks) for the following several reasons: (i) the minutiae branch and the texture branch share a number of parameters (the Inception v4 stem), reducing the model complexity that two separate models would necessitate; (ii) the spatial transformer is optimized based on both branches (i.e. learned alignment benefits both the texture-based and minutiae-based representations) avoiding two separate spatial transformer modules.

We constructed a custom architecture $f_w(x)$ with parameters $w$ to accommodate the multitask learning objective. Let $s_{w_3}(x)$ be the shared stem of our architecture and $m_{w_2}(x)$ and $t_{w_1}(x)$ be the minutiae and texture branches, respectively where $w = \{w_1, w_2, w_3\}$. The stem $s_{w_3}(x)$ provides a nonlinear mapping from the input fingerprint $I_t$ to a shared feature map $F_{map}$. The first branch $t_{w_1}(x)$ takes $F_{map}$ as input and outputs a feature representation $\mathbf{x_1} \in \mathbb{R}^{256}$. The second branch $m_{w_2}(x)$ again takes $F_{map}$ as input and outputs both a feature representation $\mathbf{x_2} \in \mathbb{R}^{256}$ and also a minutiae map $\textbf{H} \in \mathbb{R}^{128\times128\times6}$. 

Given, $\mathbf{x_1}$ and $\mathbf{x_2}$, fully connected layers are applied for identity classification logits, outputting $\mathbf{y}_1 \in \mathbb{R}^{c}$ and $\mathbf{y}_2 \in \mathbb{R}^{c}$ as follows:

\begin{equation}
\mathbf{y_i} = \mathbf{W}^{\intercal}\mathbf{x_i} + \mathbf{b},~i \in \{1, 2\}
\end{equation} where $\mathbf{W} \in \mathbb{R}^{d\times c}$, $\mathbf{b} \in \mathbb{R}^{c}$, $d$ is the number of features, and $c$ is the number of identities in the training set. Next, $\mathbf{y}_1$ and $\mathbf{y}_2$ are both passed to a softmax layer to compute the probabilities $\mathbf{\hat{y}}_1$ and $\mathbf{\hat{y}}_2$ of $\mathbf{x_1}$ and $\mathbf{x_2}$ belonging to each identity. Finally, $\mathbf{\hat{y}}_1$ and $\mathbf{\hat{y}}_2$ and their corresponding ground truth labels $y_1^g$ and $y_2^g$ can be used to compute the loss of the network from the cross-entropy loss function:

\begin{equation}
L_1(I_t, y_i^g) = -log(p(\mathbf{\hat{y}}_i = y_i^g | I_t, w_3, w_i) ~i \in \{1, 2\}
\end{equation}

For computing the loss of the minutiae map estimation side task, we employ the Mean Squared Error Loss between the estimated minutiae map $\textbf{H}$ and the ground truth minutiae map $H_g$ as follows:

\begin{equation}
L_2(I_t, H_g) = \sum_{i,j} (\textbf{H}_{i,j} - H_{g_{i,j}})^2
\end{equation} 

Finally, using the addition of all these loss terms, and a dataset comprised of $N$ training images, our model parameters $w$ are trained in accordance with:

\begin{multline}
\argmin_w \sum_{i=1}^N L_1(I_t^i, y_1^i)~+~\sum_{i=1}^N L_1(I_t^i, y_2^i)~+~ \\
\sum_{i=1}^N L_2(I_t^i,H_g^i)
\end{multline} Note, during the training, we augment our dataset with random rotations, translations, brightness, and cropping. We use the RMSProp optimizer with a batch size of 30. Regularization included dropout with a keep probability of $0.8$ and weight decay of $0.00004$. 

After the multitask architecture has converged, a fixed length feature representation can be acquired by extracting the fully connected layer before the softmax layers in both of the network's branches. Let $\mathbf{x_1} \in \mathbb{R}^{256}$ be the texture representation and $\mathbf{x_2} \in \mathbb{R}^{256}$ be the minutiae representation. Then, the final feature representation is obtained by concatenation $\mathbf{x_1}$ and $\mathbf{x_2}$ into $\mathbf{x_3} \in \mathbb{R}^{512}$ followed by normalization of $\mathbf{x_3}$ to unit length. 

\subsection{Matching}

Given two 512-dimensional, unit length representations $\mathbf{t_1}$ and $\mathbf{t_2}$, a match score $s$ is computed as the cosine similarity between the two representations. In particular:

\begin{equation}
s = \mathbf{t_1}^{\intercal} \cdot \mathbf{t_2}
\end{equation}

\subsection{Model Size Reduction}

Because our multitask architecture is very large (305 MB), we use a variant of the teacher student training schema~\cite{student} to significantly reduce the model size to 91.3 MB (with minimal loss in accuracy) such that it can be run on small embedded devices. For a student architecture, we selected the Inception v3 architecture~\cite{inceptionv3} due to its similarity to the Inception v4 architecture used as our backbone. We also experimented with smaller models for the student such as the MobileNet v2 architecture~\cite{mobilenets}, but it was unable to adequately replicate the features output by the teacher model.

The goal of the student model is to replicate the behavior of the teacher. In particular, given $N$ training images, we extract their 512-dimensional unit length feature representations $R = \{x_1, ...,x_n\}$, $n = 1...N$ from the teacher model. Then, the student model accepts the same $N$ training images, and learns to estimate the representations $R$ extracted by the teacher via the L2-loss function. Let $student_w(I_n) = x_n^s$ be the non-linear mapping of the student model from the input images $I_n$ to the fingerprint representations $x_n^s$ where $x_n^t$ is the ground truth representation extracted by the teacher. Then, the student is trained by optimizing the L2-loss function:

\begin{equation}
\argmin_w \sum_{i=1}^N L(x_n^s, x_n^t)
\end{equation} where

\begin{equation}
L(x_n^s, x_n^t) = \frac{1}{2}||x_n^t - x_n^s||_2^2
\end{equation}

\section{Experimental Results}

The baseline experimental results are comprised of two state-of-the-art commercial minutiae-based matchers~\footnote{COTS used were Verifinger SDK v6.0 and Innovatrics~\cite{innovatrics}, a top performer in NIST FpVTE 2012.} and the Inception v4 model trained in a manner similar to~\cite{index1} (i.e. no alignment or domain knowledge is added to the network). Our proposed method includes three major steps: (i) adding a spatial transformer (Inception v4 + STN) for alignment, (ii) adding domain knowledge via multitask learning (Inception v4 + STN + MTL), and (iii) model size reduction via a smaller student model (Student). Recognition accuracy for all of these methods are reported in the following subsections.

\subsection{Testing Datasets}

\begin{figure}[h]
  \centering
  \subfloat[NIST SD4 rolled fingerprint]{\includegraphics[scale=0.25]{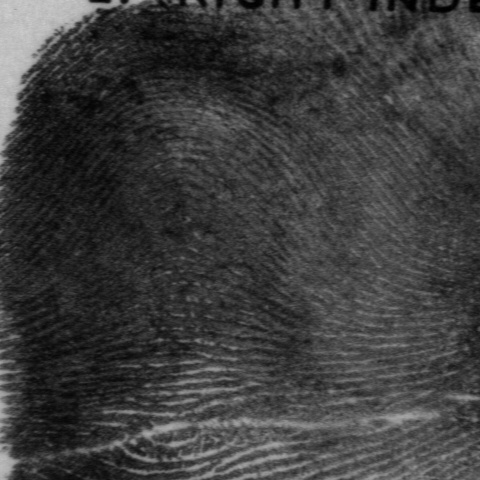}\label{fig:f12}}
  \hfill
  \subfloat[FVC 2004 plain fingerprint]{\includegraphics[scale=0.25]{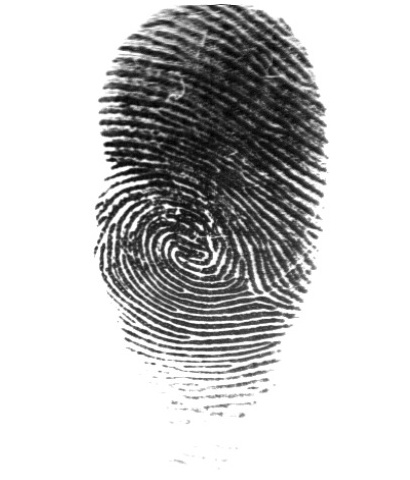}\label{fig:f22}}
  \caption{Examples from benchmark datasets. (a) A smudgy rolled fingerprint from NIST SD4 and (b) a distorted plain (slap) fingerprint from FVC 2004 DB1.}
  \label{fig:db}
\end{figure}

We use the NIST SD4 ~\cite{sd4} and FVC 2004 DB1~\cite{fvc2004} databases for benchmarking our fixed-length matcher. The NIST SD4 dataset is comprised of 4000 \textit{rolled} fingerprints (2000 unique fingers with 2 impressions per finger) (Fig.~\ref{fig:f12}) and the FVC 2004 DB1 is comprised of 800 \textit{plain} fingerprints (100 unique fingers with 8 impressions per finger) (Fig.~\ref{fig:f22}). Both of these datasets are considered challenging datasets, even for commercial matchers. 

\begin{table*}[h!]
 \centering
 \caption{Verification Results (1-to-1 comparison)}
 \vspace{-0.75em}
\begin{threeparttable}
\begin{tabular}{|c||c|c|c|c|c|c|}
 \hline
 \specialcell{Database // Method } & \specialcell{Inception v4 \\(unmodified)} & COTS A & COTS B & \specialcell{Inception v4 \\(+STN)} & \specialcell{Inception v4 \\(+STN + MTL)} & Student\\
 \hline
  FVC 2004\tnote{1} & 58.4\% & 92.3\% & \textbf{95.6\%} & 65.1\% & 93.2\% & 86.4\% \\
 \hline
 \hline
 NIST SD4\tnote{2} & 92.0\% & 96.1\% & \textbf{97.0\%} & 94.0\% & 96.9\% & 95.8\% \\
 \hline
\end{tabular}
\begin{tablenotes}
\item[1] \small{TAR @ FAR = 0.1\% is reported due to the insufficient number of imposters for FAR = 0.01\%}
\item[2] \small{TAR @ FAR = 0.01\% is reported.}
\end{tablenotes}
\end{threeparttable}
\label{table:ver}
\vspace{-0.5em}
\end{table*}

\begin{table*}[h!]
 \centering
 \caption{NIST SD4 Rank-1 Search Results (background 2000 fingerprints)}
 \vspace{-0.75em}
\begin{threeparttable}
\begin{tabular}{|c|c|c|c|c|c|}
 \hline
 \specialcell{Inception v4 \\(unmodified)} & COTS A & COTS B & \specialcell{Inception v4 \\(+STN)} & \specialcell{Inception v4 \\(+STN + MTL)} & Student\\
 \hline
 94.8\% & 96.4\% & 97.3\% & 97.0\% & \textbf{97.9\%} & 96.8\% \\
 \hline
\end{tabular}
\end{threeparttable}
\label{table:search}
\vspace{-1.0em}
\end{table*}

\subsection{Recognition Accuracy}

From the results (Tables~\ref{table:ver}  and~\ref{table:search}), we can observe that the unmodified off-the-shelf Inception v4 model performs worse than COTS matchers. Adding the spatial transformer module to the Inception v4 backbone (Inception v4 + STN) boosts the accuracy in all testing scenarios. With added domain knowledge via our proposed multitask learning architecture, accuracy is further boosted and becomes competitive with the two COTS matchers. 

We posit the reason for the low performance of the Inception v4 model without fingerprint domain knowledge on the FVC 2004 database is because the fingerprint image characteristics in FVC 2004 are much different than the characteristics of our training data. In particular, our longitudinal training database~\cite{longitudinal} is comprised of large, inked, and scanned rolled fingerprints similar to NIST SD4 (Fig.~\ref{fig:f12}). Meanwhile, FVC consists of small area fingerprints acquired with a fingerprint reader (Fig.~\ref{fig:f22}). Therefore, the Inception v4 model without domain knowledge ``overfits" to rolled ink fingerprints. As can be seen in Table~\ref{table:ver}, our approach for adding domain knowledge via the minutiae map in combination with our multitask architecture significantly improves the generalizability of the deep fingerprint representations on testing datasets different from the rolled fingerprints in the training dataset (increasing the TAR from 65.1\% to 93.2\%).

We also note that our use of the teacher student training paradigm enabled us to significantly reduce our model size with only a small loss in accuracy.  

\subsection{Speed}

Extracting fixed length representations enables us to perform matching \textbf{several orders of magnitude} faster than minutiae-based matchers (Table~\ref{table:speed}). This has tremendous benefits for large scale searches against millions or even billions of fingerprints in the background. Admittedly our representation extraction time is slightly longer than minutiae extraction. However, representation extraction is only performed once whereas matching is performed many times.

\begin{table}[h]
\small
 \centering
 \caption{Speed Comparison\tnote{1}}
 \vspace{-0.75em}
\begin{threeparttable}
\begin{tabular}{|c|c|c|}
 \hline
 Matcher & \specialcell{Representation \\ Extraction (ms)} & \specialcell{Matches / \\ Second} \\
 \hline
 \specialcell{COTS A} & 142 & 22 \\
  \hline
 \specialcell{COTS B} & \textbf{113} & 20 \\
  \hline
   \specialcell{Inception v4 (+ STN + MTL)} & 313 & \textbf{682,594} \\
  \hline
  Student & 157 & \textbf{682,594} \\
  \hline
\end{tabular}
\begin{tablenotes}
\item[1] Computed on an i5 3.3 GHz processor with 8 GB of RAM.
\end{tablenotes}
\end{threeparttable}
\label{table:speed}
\vspace{-1.0em}
\end{table}

\subsection{Large Scale Search}

Having benchmarked our matcher against two minutiae-based COTS matchers, we further demonstrate its efficacy for large scale fingerprint search by computing the identification accuracy on NIST SD4 with a supplemented background (gallery) of 282,000 unique rolled fingerprints from a longitudinal database~\cite{longitudinal}. With a background of 282,000 fingerprint images, we achieve a Rank-1 Identification accuracy of 94.3\%, only a slight drop from the search performance with a 2K background.  A single search of a probe against the background takes only 416 milliseconds with our fixed-length matcher. Due to the limited speed of the COTS SDKs, we are unable to perform large scale retrieval experiments with them.

\section{Summary}

We have introduced a custom deep network which uses built-in alignment and embedded domain knowledge to extract discriminative fixed-length fingerprint representations. The strong advantages of our fixed-length representations over prevailing variable length minutiae templates are: 

\begin{enumerate}
\item Orders of magnitude faster fingerprint matching (682,594 matches / second vs. 22 matches / second), invaluable for large scale search.

\item The ability to provide strong encryption (a significant challenge with variable length templates).

\item Additional discriminative textural cues beyond just minutiae, enabling competitive recognition accuracy with minutiae-based matchers.
\end{enumerate}
 
{\small
\bibliographystyle{ieeetr}
\bibliography{egbib}
}

\end{document}